%% file: acl_latex.tex
\pdfoutput=1

\documentclass[11pt]{article}

\usepackage[preprint]{acl}

\usepackage{times}
\usepackage{latexsym}

\usepackage[T1]{fontenc}

\usepackage[utf8]{inputenc}

\usepackage{microtype}

\usepackage{inconsolata}

\usepackage{graphicx}
\usepackage{booktabs}
\usepackage{multirow}
\usepackage{colortbl}
\usepackage{tabularx}

\usepackage{url}

\title{Is Fine-Tuning an Effective Solution? Reassessing Knowledge Editing for Unstructured Data}

\author{
 \textbf{Hao Xiong}\thanks{Equal contribution},
 \textbf{Chuanyuan Tan}\textsuperscript{*},
 \textbf{Wenliang Chen}\thanks{Corresponding author},
\\
 School of Computer Science and Technology, Soochow University, China,
\\
 \small{
   \texttt{\{hxiongxionghao, cytan17726\}@stu.suda.edu.cn}, \texttt{wlchen@suda.edu.cn}
 }
}

\begin{document}
\maketitle
\begin{abstract}
Unstructured Knowledge Editing (UKE) is crucial for updating the relevant knowledge of large language models (LLMs). It focuses on unstructured inputs, such as long or free-form texts, which are common forms of real-world knowledge. Although previous studies have proposed effective methods and tested them, some issues exist: (1) Lack of Locality evaluation for UKE, and (2) Abnormal failure of fine-tuning (FT) based methods for UKE. To address these issues, we first construct two datasets, UnKEBench-Loc and AKEW-Loc (CF), by extending two existing UKE datasets with locality test data from the unstructured and structured views. This enables a systematic evaluation of the Locality of post-edited models. Furthermore, we identify four factors that may affect the performance of FT-based methods. Based on these factors, we conduct experiments to determine how the well-performing FT-based methods should be trained for the UKE task, providing a training recipe for future research. Our experimental results indicate that the FT-based method with the optimal setting (FT-UKE) is surprisingly strong, outperforming the existing state-of-the-art (SOTA). In batch editing scenarios, FT-UKE shows strong performance as well, with its advantage over SOTA methods increasing as the batch size grows, expanding the average metric lead from +6.78\% to +10.80\%.~\footnote{Our code and data will be released on Github.}

\end{abstract}

\input{secs/1.intro.tex}

\input{secs/2.related.tex}

\input{secs/3.dataset.tex}

\input{secs/4.method.tex}

\input{secs/5.exp.tex}

\input{secs/6.conclusion.tex}

\section*{Limitations}
This paper conducts analytical experiments on several factors of FT-based methods and derives a training recipe for the UKE task. We opted for a set of experiments that researchers have proven to have competitive performance in the SKE task, rather than enumerating all possible combinations due to limitations in computational resources. Specifically, the settings we skip include: (1) full-parameter fine-tuning, which involves training all parameters of all layers rather than just a part of a layer component, and (2) other combinations for the factor \textit{Component}, such as editing joint configurations of \textit{q$_{proj}$}, \textit{k$_{proj}$}, \textit{v$_{proj}$}, \textit{o$_{proj}$}, \textit{down$_{proj}$}. Considering that the current FT-UKE in the settings we experimented with already surpasses the existing SOTA methods, we decide not to pursue further exploration of the aforementioned settings, opting to leave them for future work.

\bibliography{custom}

\appendix

\input{secs/7.appendix}

\end{document}

%% file: secs/1.intro.tex
\section{Introduction}
\begin{figure}[t]
  \includegraphics[width=\columnwidth]{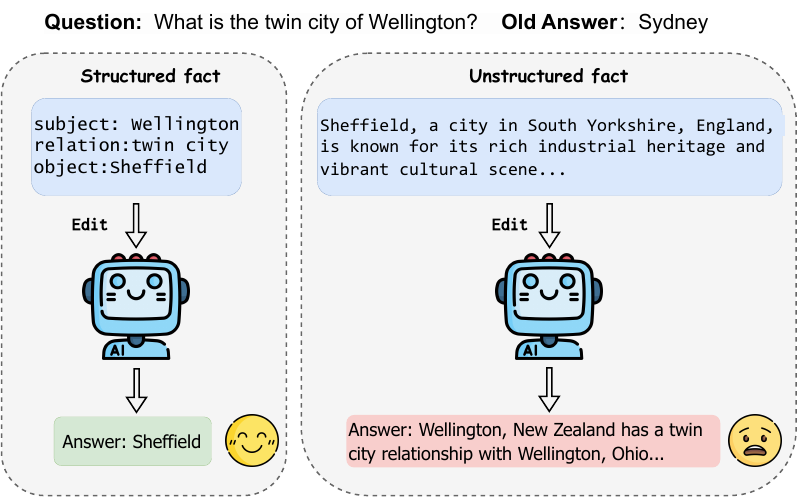}
  \caption{Comparison between structured and unstructured knowledge editing. While structured editing operates on predefined factual triples, unstructured editing involves open-text modifications, introducing greater difficulty.}
  \label{fig:intro}
\end{figure}

With the rapid development of large language models (LLMs) \cite{brown2020language, achiam2023gpt, touvron2023llama, bai2023qwen} across various domains, the ability to update model's internal knowledge, known as knowledge editing, has gained increasing attention~\citep{ meng2022rome, yao-etal-2023-editing, zhang2024comprehensive}. The goal of knowledge editing is to accurately update specific knowledge within a model while minimizing the impact on other unrelated knowledge. 

Substantial research focuses on Structured Knowledge Editing (\textbf{SKE}) \citep{meng2022rome, hu-etal-2024-wilke, fang2024alphaedit}: editing knowledge represented as triples (subject, relation, object). To evaluate the effectiveness of these SKE methods, researchers have developed dedicated datasets and conducted evaluations from three perspectives: (1) \textbf{Edit Success}: correctly learns the new knowledge, (2) \textbf{Generalization}: generalizes it to paraphrased or rephrased queries, and (3) \textbf{Locality}: preserves performance on unedited knowledge.

As the task of SKE has achieved significant success, researchers are increasingly focusing on Unstructured Knowledge Editing (\textbf{UKE}) \cite{wu2024akew, deng2024unke, jiang2025anyedit}. This task aims to modify knowledge embedded in long or free-form text. 
As shown in Figure~\ref{fig:intro}, unlike structured knowledge represented as triples, unstructured knowledge appears in the form of extended text, containing rich information and complex contextual dependencies. 

Although the researchers have proposed effective UKE methods and validated their effectiveness on UKE datasets, our preliminary investigations and experiments reveal the following issues:
(1) \textbf{Lack of Locality evaluation for UKE}: Existing UKE datasets are primarily designed to evaluate two aspects: Edit Success and Generalization. However, they lack datasets specifically tailored for assessing Locality. Instead, they rely solely on the general assessment dataset MMLU~\citep{hendrycks2020measuring} for this purpose. In terms of results, this evaluation lacks differentiation, with the gap between the worst method and pre-edit results not exceeding ±1.2\%~\citep{deng2024unke}.
(2) \textbf{Abnormal failure of fine-tuning (FT) based methods for UKE}: While FT-based methods serve as important baselines and are competitive in the SKE task, they reportedly underperform in the UKE task. Even in terms of the Edit Success metric, where FT-based methods can surpass specially designed SKE methods, they still do not perform well in the UKE task. We argue that this is an abnormal phenomenon, which requires systematic experiments and analysis to identify the reasons or to determine if there is a misunderstanding.

To address these issues, we first construct two new datasets, \textbf{UnKEBench-Loc} and \textbf{AKEW-Loc (CF)}, by extending UKE datasets \textbf{UnKEBench}~\citep{deng2024unke} and \textbf{AKEW} (\textbf{C}ounter\textbf{F}act)~\citep{wu2024akew}. This extension involves incorporating three types of Locality test data. Specifically, we sample two types of unstructured data and one type of structured data.

Furthermore, we identify four factors that influence the performance of FT-based methods in knowledge editing from previous studies \citep{zhu2020modifying,zhang2024comprehensive,hu2022lora}. These factors are frequently discussed in previous SKE research~\citep{meng2022rome,zhang2024comprehensive,li2024pmet}: (1) \textbf{Loss Calculation Scope}: choosing final prediction token or all target tokens to calculate loss;
(2) \textbf{Layer Selection}: deciding whether to edit a single layer or all layers of the target model; 
(3) \textbf{Component Selection}: for the selected layer(s), determining whether to edit the feed-forward network or the attention projections;
(4) \textbf{Chat Template}: deciding whether to adopt a chat template for the target model. 
Through experimental analysis, we identify the optimal settings for each factor in the UKE task, which can benefit future research.

In summary, our contributions are as follows:
\begin{itemize}
    \item We construct two UKE datasets, UnKEBench-Loc and AKEW-Loc(CF), to directly and comprehensively evaluate UKE Locality. These datasets include a total of 5,925 Locality test data across three types: two types of unstructured data and one type of structured data. To our best knowledge, these expanded datasets are the first UKE datasets containing multi-type, well-designed test data that support Locality evaluation for UKE task.
    \item We outline the factors influencing the performance of FT-based methods. Through detailed experimental analysis, we provide a training recipe for FT-based methods in the UKE task, which offers a strong training setup for future research.
    \item Based on evaluation, we find that the FT-based method with the optimal setting (FT-UKE) is surprisingly strong, surpassing all the SOTA methods. We further explore the performance of UKE methods in the batch editing scenarios. Surprisingly, FT-UKE maintains its advantage over SOTA methods, with a larger average increase from +6.78\% to +10.80\%.
    
\end{itemize}

%% file: secs/2.related.tex
\section{Related Work}
\subsection{Knowledge Editing}

Research on Structured Knowledge Editing (SKE) is well-developed and can be categorized into three main approaches: locate-and-edit~\citep{meng2022rome, meng2023memit, fang2024alphaedit}, meta-learning~\citep{mitchellfast, tan2024massive}, and retrieval-based methods~\citep{zheng-etal-2023-edit,wang2024wise}. For the UKE task, current methods primarily follow the locate-and-edit approach, such us UnKE~\citep{deng2024unke} and AnyEdit~\citep{jiang2025anyedit}. These methods enhance their ability to handle unstructured knowledge by updating all parameters within a single transformer layer. Since most existing UKE methods adopt the locate-and-edit approach, we select SKE baseline methods for comparison that also focus on this approach.

Besides, knowledge editing can be categorized in two scenarios by the number of data points edited per test: single editing and batch editing~\citep{meng2023memit}. Single editing involves testing after editing each individual piece of knowledge. In contrast, batch editing refers to editing \( n \) pieces of data at once, where \( n \) is called "batch size". Both single and batch editing have been extensively discussed in SKE task. However, in the UKE task, research primarily focuses on the effectiveness of single editing, with only a few studies reporting performance in batch editing scenarios~\citep{deng2024unke}.

\subsection{Evaluation Settings for Knowledge Editing}

In the SKE task, researchers calculate metrics by assessing the consistency between the post-edit model's output and the expected output. Specifically, for Edit Success and Generalization, the expected output is the edited knowledge; for the Locality test, the expected output is the pre-edit model's output~\citep{zhang2024comprehensive}. Due to the limited length of structured knowledge, consistency is typically calculated at the token level. For the UKE task, token-level calculations are unsuitable due to text length. \citet{deng2024unke} introduces a method based on BERT Score \citep{zhang2019bertscore} and ROUGE-L \citep{lin-2004-rouge} to evaluate the semantic and lexical similarity for UKE Edit Success and UKE Generalization. We apply this method for UKE Locality calculation as well. Although previous research does not specifically design Locality test data, \citet{deng2024unke} samples data from MMLU~\citep{hendrycks2020measuring}, testing a few multiple-choice questions after a single edit. They calculate the change in accuracy before and after editing, which reflects Locality. However, the accuracy for all methods shows only minor differences from the pre-editing performance, reportedly not exceeding ±1.2\% when editing Llama2-7B-Chat. This raises concerns about whether this dataset can effectively differentiate between different methods, especially those with similar capacities. This underscores the need to construct specialized localization data that is better suited for UKE tasks.

%% file: secs/3.dataset.tex
\begin{figure}[t]
  \includegraphics[width=\columnwidth]{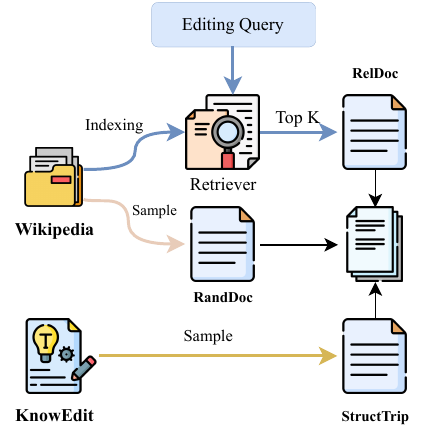}
  \caption{Data Collection Process. We sample unstructured data (\textbf{RelDoc}, \textbf{RandDoc}) from Wikipedia and structured data (\textbf{StructTrip}) from structured knowledge editing dataset KnowEdit.}
  \label{fig:datasets}
\end{figure}
\section{Datasets for Locality Test}
\label{sec:dataset}

In this section, we introduce how we expand UKE datasets with Locality data from the unstructured and structured views.
As shown in Figure~\ref{fig:datasets}, we sample two types of unstructured data from Wikipedia and one type of structured data from KnowEdit~\citep{zhang2024comprehensive}, a structured knowledge editing dataset. Pre-process details are listed in Appendix~\ref{app:dataset}.

\begin{itemize}
    \item \textbf{Relevant unstructured data (RelDoc)}: For each editing query, we retrieve a Wikipedia document that is semantically related but factually disjoint from the target knowledge. To facilitate effective document retrieval, we train a Dense Passage Retrieval (DPR)~\citep{karpukhin2020dense} model using a collection of question-answering datasets. The training setup is detailed in Appendix~\ref{app:dataset}. To ensure factual disjointness, we exclude documents containing same entity-relation pairs as those appearing in the editing query~\footnote{entity-relation pairs are extracted by~\href{https://github.com/philipperemy/stanford-openie-python}{OpenIE}.}. Using RelDoc, we can assess the influence of editing methods on semantically related unstructured knowledge.

    \item \textbf{Random unstructured data (RandDoc)}: We randomly sample a Wikipedia document, excluding the top 100 documents most relevant to the editing query. This type of data allows us to investigate global Locality by assessing the post-editing impact on distant and unrelated unstructured knowledge.
    
    \item \textbf{Structured data (StructTrip)}: We also sample structured knowledge from the SKE dataset KnowEdit, which is unrelated to the edited unstructured knowledge. This enables us to evaluate how the editing of unstructured knowledge impacts unrelated structured knowledge.
  
\end{itemize}

Based on the above approach, we expand two UKE datasets \textbf{UnKEBench}~\citep{deng2024unke} and \textbf{AKEW}(\textbf{C}ounter\textbf{F}act), a subset of AKEW~\citep{wu2024akew}. We exclude the other two AKEW subsets (MQUAKE-CF, WikiUpdate) because they lack the data necessary for evaluating Generalization, rendering them less suitable for a comprehensive assessment. We mark the expanded datasets as \textbf{UnKEBench-Loc} and \textbf{AKEW-Loc (CF)} respectively, and summarize the statistics of them in Table~\ref{tab:dataset_stats}. 

\input{table/dataset_stats}

%% file: table/dataset_stats.tex
\begin{table}[t]
\centering
\resizebox{0.48\textwidth}{!}{
\begin{tabular}{lcc}
\toprule
 & \textbf{UnKEBench-Loc} & \textbf{AKEW-Loc (CF)} \\
\midrule
\# Editing query & 1,000 & 975 \\
\multicolumn{3}{l}{\# Locality test data} \\
\hspace{1em}Total & 3,000 & 2,925\\
\hspace{1em}RelDoc & 1,000 & 975 \\
\hspace{1em}RandDoc & 1,000 & 975\\
\hspace{1em}StructTrip & 1,000 & 975\\
\midrule
\multicolumn{3}{l}{Average length of Locality test data*}\\
\hspace{1em}RelDoc & 118.56 & 117.43 \\
\hspace{1em}RandDoc & 116.19 & 116.31\\
\hspace{1em}StructTrip & 8.34 & 8.18\\
\bottomrule
\end{tabular}
}
\caption{Statistics of UnKEBench-Loc and AKEW-Loc (CF), including the number of editing query, and the number and average length of Locality test data. *: Calculated by NLTK~\citep{bird-loper-2004-nltk}.}
\label{tab:dataset_stats}
\end{table}

%% file: secs/4.method.tex
\section{Revisiting Fine-tuning based method for UKE}
\label{sec:method}

In this section, we revisit fine-tuning (FT) based approaches for UKE, including: (1) direct weight fine-tuning, which directly updates the original model weights (e.g., FT-L \citep{zhu2020modifying} and FT-M~\citep{zhang2024comprehensive}); and (2) additional parameter fine-tuning, which introduces additional trainable modules, such as adapters, while keeping the original model weights frozen (e.g., AdaLoRA~\citep{hu2022lora}).
Although frequently adopted in prior studies~\citep{meng2022rome,deng2024unke}, these methods show suboptimal performance for UKE. We argue that this underperformance is abnormal and does not stem from the fundamental limitations of fine-tuning itself. Therefore, we conduct a systematic analysis of FT-based methods considering four factors: loss calculation scope, layer selection, component selection, and chat template. Table~\ref{tab:baseline-summary} lists the choices for these factors.

\paragraph{Loss Calculation Scope} The scope of loss calculation is crucial for aligning training signals with the desired output. One approach, used by FT-L, involves calculating the loss solely on the final prediction token to maximize the probability of the output. In contrast, another approach, employed by FT-M and AdaLoRA, calculates the loss across all tokens of the output.

\paragraph{Layer Selection} The choice of which layer to edit is a critical factor in knowledge editing. \citet{meng2022rome} introduced a causal tracing technique to identify the most causally relevant layers for intervention, subsequent work has shown that the selected layer can significantly impact editing outcomes. Based on these insights, we explore two strategies: editing a single middle layer or updating all transformer layers. These design choices are informed by empirical findings from frameworks such as EasyEdit \citep{wang2023easyedit}.

\paragraph{Component Selection} Direct weight fine-tuning methods directly modify the weights of specific components in the original model, often targeting the feed-forward network (FFN) layers, such as \textit{down$_{proj}$} in the MLP. In contrast, additional parameter fine-tuning methods, such as AdaLoRA, introduce low-rank adapter modules into the attention projections (e.g., \textit{q$_{proj}$, k$_{proj}$, v$_{proj}$, o$_{proj}$}), allowing for efficient adaptation while keeping the base model frozen. These designs are not strictly exclusive. To better understand how different editable components affect editing performance, we follow prior work \citep{zhang2023adalora,wang2023easyedit} and evaluate several common configurations under both paradigms. A summary of these configurations is provided in Appendix~\ref{app:configuration}. 

\paragraph{Chat Template} Unstructured knowledge editing typically involves natural language instructions as inputs. For instruction-tuned language models, the use of standardized chat templates helps align the input format with the model’s pretraining and fine-tuning distribution. In contrast, editing without such templates may introduce discrepancies between the input and the model’s expectations, potentially reducing editing effectiveness. In our analysis, we compare variants with and without standardized chat templates to examine their impact on editing performance.

By reevaluating fine-tuning based methods with refined configurations and instruct-compatible settings, we aim to establish a strong setup for FT-based methods in unstructured knowledge editing and provide a training recipe for future research.

\begin{table}[t]
\centering
\small
\resizebox{0.48\textwidth}{!}{
\begin{tabular}{ll}
    \toprule
    \textbf{Factor} & \textbf{Choices} \\
    \midrule
        \multirow{2}{*}{Loss Calculation Scope} & final prediction token \\
        & all target tokens \\
    \midrule
        \multirow{2}{*}{Layer Selection} & single layer \\
        & all layers \\
    \midrule
        \multirow{2}{*}{Component Selection*} & FFN \\
        & Attention\\
    \midrule
        \multirow{2}{*}{Chat Template} & w. template \\
        & w/o. template\\
    \bottomrule
\end{tabular}
}
\caption{Factors we considered for the FT-based method in UKE task. *: Detailed settings are provided in Appendix \ref{app:configuration}.}
\label{tab:baseline-summary}
\end{table}

%% file: secs/5.exp.tex
\section{Experiments}
In this section, we conduct experiments on datasets introduced in \S~\ref{sec:dataset}: UnKEBench-Loc and AKEW-Loc (CF).
\subsection{Experiment Setup}

\paragraph{Language Models} Following~\citet{jiang2025anyedit}, we use Llama3-8B-Instruct~\citep{llama3modelcard} and Qwen2.5-7B-Instruct~\citep{qwen2.5} as the language models to be edited.
\paragraph{Baseline Methods}
We report two UKE methods: UnKE~\citep{deng2024unke} and AnyEdit~\citep{jiang2025anyedit}, as well as our adopted FT-based methods FT-UKE and AdaLoRA-UKE, which are best-performing settings across all settings we discussed in \S~\ref{sec:method}. Additionally, we report three widely used SKE methods for comparison: ROME~\citep{meng2022rome}, MEMIT~\citep{meng2023memit}, and AlphaEdit~\citep{fang2024alphaedit}. To demonstrate the difference in performance before and after editing, we also report the performance before editing, denoted as Pre-edit. Details of baseline methods are listed in Appendix~\ref{app:exp_details}.

\paragraph{Evaluation Metrics}

We evaluate from three perspectives: 
(1) Edit Success (\textbf{Ori}): Tests whether the model correctly answers the original edit query with the new target.
(2) Generalization (\textbf{Para}): Uses paraphrased queries to assess whether the edit generalizes beyond the original phrasing.
(3) Locality (\textbf{Loc}): Measures whether unrelated knowledge is preserved by checking if the model's output on unaffected inputs remains unchanged.
Finally, we report the average of Ori, Para, and Loc as a comprehensive metric, Overall score (\textbf{OA}).

Following~\citet{jiang2025anyedit, deng2024unke}, we use two metrics to measure the similarity between post-edited model’s output and reference output: BERT Score (\textbf{BS})~\citep{zhang2019bertscore} for semantic similarity and ROUGE-L (\textbf{RL})~\citep{lin-2004-rouge} for lexical similarity.~\footnote{Specifically, we employ \href{https://huggingface.co/sentence-transformers/all-MiniLM-L6-v2}{all-MiniLM-L6-v2} to compute BERT Score.}

\subsection{Main Result}
\label{sec:main_res}

\input{table/main_res}

We compare FT-based methods with other strong UKE and SKE methods for editing two LLMs. 
Consistent with~\citet{deng2024unke}, we adopt a batch size of 1 and set the decoding temperature to 0.001. The main results are listed in Table~\ref{tab:main_res}. 
According to these results, we have the following observations:

(1) \textbf{The best FT-based method, FT-UKE, consistently outperforms the SOTA UKE methods}. We surprisingly find that FT-UKE outperforms all methods, including the SOTA UKE methods, except in one instance (BS of OA in AKEW-Loc (CF), editing Llama3). Compared to the best UKE method in the previous studies, AnyEdit, it exceeds by 4.44\% and 9.12\% in the BS and RL of OA score on AKEW-Loc (CF), with an average advantage of 6.78\%. 
Notably, even when compared to the results reported in the original paper, FT-UKE still demonstrates a significant advantage.
Besides, AdaLoRA-UKE also demonstrated very competitive performance. For example, in the experiments on Qwen2.5-7B-Instruct, its OA consistently surpassed the SOTA UKE methods.

(2) \textbf{The failure of FT-based methods in the previous studies may be attributed to the use of suboptimal settings}. Comparing FT-based methods reported by the previous studies and FT-UKE, we find that while FT-based methods can achieve strong performance, they require careful selection of important factors. Therefore, we encourage future researchers to adopt our training recipe to build a strong baseline for UKE.

(3) \textbf{UnKE and AnyEdit are still strong methods that significantly outperform the SKE methods}. Taking the results of Llama3-8B-Instruct as an example, UnKE and AnyEdit demonstrate a significant advantage over the best SKE method, ROME, across all datasets. For instance, UnKE's BS and RL of OA on UnKEBench-Loc are around 10\% and 30\% higher than ROME's. We observe similar trends in other settings. This suggests that although UnKE and AnyEdit are not as powerful as FT-UKE, they remain competitive methods for the UKE task.

\subsection{Analysis of Factors for FT-based Methods}
\label{sec:dif_ft}

\input{table/dif_lora_ft}

In this section, we edit Llama3-8B-Instruct on AKEW-Loc (CF) using FT-based methods, applying the factor settings discussed in \S~\ref{sec:method}. As shown in Table~\ref{tab:dif_lora_ft}, our analysis yields the following key findings:

(1) \textbf{Calculate loss on \textit{final prediction token} is not a good choice for UKE}. We find settings that calculate loss using only the \textit{final prediction token} underperform those using \textit{all target tokens} by over 50\% in terms of OA. This significant difference indicates that using the \textit{final prediction token} is not a good choice for the \textit{loss calculation scope} in the UKE task.

(2) \textbf{The optimal choice for \textit{component} may differ for FT-based methods between SKE and UKE tasks, while the choice for \textit{layer} and \textit{chat template} remains the same}. The best settings for additional parameter fine-tuning (AdaLoRA-UKE) and direct weight fine-tuning (FT-UKE) are highlighted in green in the table. For the optimal choice for \textit{component} in UKE task, AdaLoRA-UKE involves whole attention projections (\textit{q$_{proj}$}, \textit{k$_{proj}$}, \textit{v$_{proj}$}, \textit{o$_{proj}$}), which differs from that in SKE (\textit{q$_{proj}$}, \textit{v$_{proj}$})~\cite{wang2023easyedit}. As for FT-UKE, the optimal choice remains the same (\textit{down$_{proj}$}) between SKE and UKE. Similarly, the optimal choice of \textit{layer} remains consistent, with \textit{all} for AdaLoRA-UKE, and \textit{one} for FT-UKE.
As for \textit{chat template}, applying it during editing significantly boosts performance across all settings. Detailed comparisons can be found in Appendix~\ref{app:configuration}.

\subsection{Performance in Batch Editing Scenarios}

\begin{figure*}[t]
    \centering
    \includegraphics[width=0.86\textwidth]{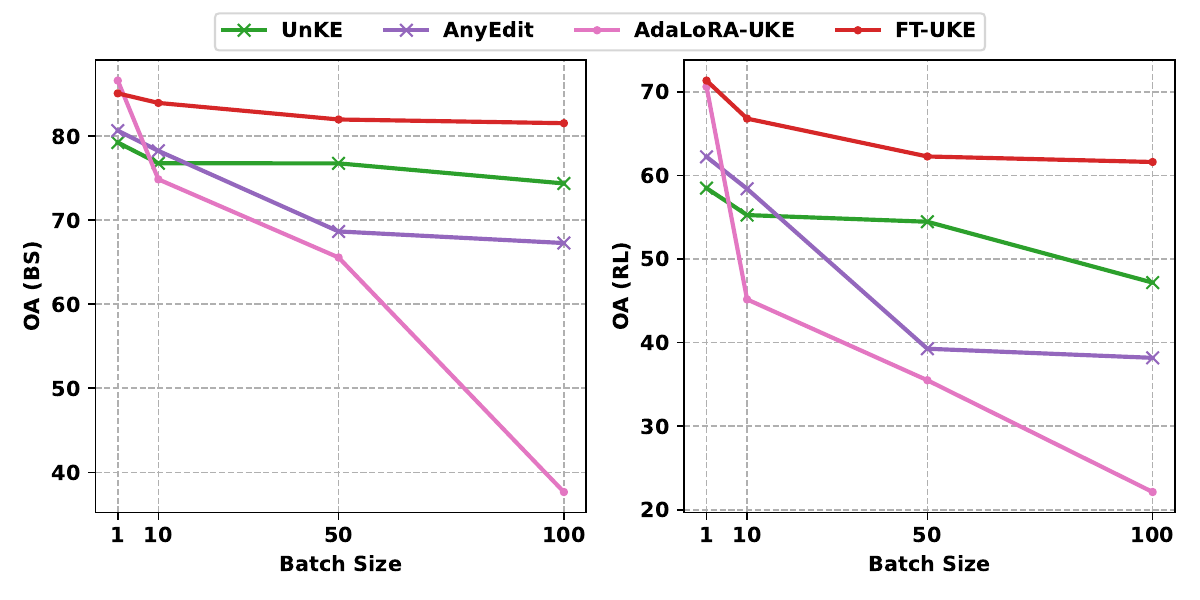}
    \caption{OA of different methods for editing Llama3-8B-Instruct on AKEW-Loc (CF) in batch editing scenarios. FT-UKE has advantages over SOTA UKE methods across different batch sizes, and the magnitude of these advantages increases with larger batch sizes. Detailed results are listed in Appendix (Table~\ref{tab:batch_res}).}
    \label{fig:batch_res}
\end{figure*}

To further assess the robustness of various editing methods under batch editing scenarios, we edit Llama3-8B-Instruct using different batch sizes (1, 10, 50, and 100) on the AKEW-Loc (CF) dataset. 
We aim to investigate how FT-UKE and AdaLoRA-UKE perform in batch editing scenarios.
Therefore, we present the results of four methods in Figure~\ref{fig:batch_res}: FT-UKE, AdaLoRA-UKE, and two UKE methods, UnKE and AnyEdit. These methods perform well in the single editing scenario (\S~\ref{sec:main_res}).

As shown in Figure~\ref{fig:batch_res}, \textbf{FT-UKE maintains its advantage over SOTA UKE methods in batch editing scenarios}, demonstrating strong robustness and effectiveness. All methods exhibit a general decline in performance as batch size increases. However, FT-UKE degrades more gradually, which results in a progressively larger advantage in average OA over other methods, increasing from 6.78\% to 10.80\%. In contrast, \textbf{AdaLoRA-UKE suffers the steepest drop}, indicating greater sensitivity to batch interference. 
Specifically, AdaLoRA-UKE shows a more significant decline compared to other methods as the batch size increases to 10, particularly in the OA of RL, where it decreases from 70.60\% to 45.16\%. When the batch size reaches 100, AdaLoRA-UKE becomes almost ineffective, with an OA of only 37.66/22.12 (BS/RL). As for UnKE and AnyEdit, although AnyEdit is the SOTA UKE method in single editing scenarios with a batch size of 1, it is surpassed by UnKE when the batch size increases to 50. 

For future work, we recommend incorporating batch editing scenarios into testing to more comprehensively evaluate the effectiveness of UKE methods. Additionally, FT-UKE is the most suitable FT-based method for comparison, rather than AdaLoRA-UKE, which may not perform well with large batch sizes.

\subsection{Comparison with Locality Evaluation Using General Assessment Dataset}

\input{table/mmlu_loc}

Previous studies rely on a general assessment dataset MMLU to evaluate the Locality of UKE \citep{deng2024unke}, by observing changes in multiple-choice accuracy before and after editing. However, \textbf{we argue that such evaluations are insufficient for Locality evaluation}. 
To support our argument, we utilize the Locality data constructed by \citet{deng2024unke} based on MMLU, referred to as MMLU-Loc, instead of the Locality data we constructed, to report the performance on datasets UnKEBench-Loc and AKEW-Loc (CF) for editing Llama3-8B-Instruct.

As shown in Table~\ref{tab:mmlu_loc}, \textbf{most editing methods exhibit performance similar to the pre-edit model on MMLU-Loc.} Even the method with the lowest accuracy (ROME) shows a decline of only 1.17\% from the pre-edit model.
Given that MMLU-Loc provides only a few multiple-choice questions for the Locality data of a single edit query, and considering that evaluations based on such a small set can be random, we are concerned that this narrow gap may fail to accurately reflect the differences in Locality between methods, especially when the capabilities of them are similar. For instance, the differences between AnyEdit and UnKE are very small, less than 0.1\% on AKEW-Loc (CF).

\textbf{In contrast, the Locality data we constructed reveals clearer distinctions.} Taking the BS score of AKEW-Loc (CF) as an example: (1) The minimum gap between methods and pre-edit is 16.56\%, which is much larger than that on MMLU-Loc (1.17\%); (2) The difference between UnKE and AnyEdit is 1.42\%, which is significantly larger than that on MMLU-Loc as well.
This demonstrates that our datasets offer a more sensitive and informative assessment of Locality. This is attributed to (1) Our data consists of three types, including both structured and unstructured data, and is meticulously designed for edit queries. (2) Our evaluation framework is similar to SKE datasets by comparing the consistency between the output of post-edit and pre-edit models, which is more suitable for the knowledge editing task \cite{deng2024unke, jiang2025anyedit}.

%% file: table/main_res.tex
\begin{table*}[t]

\centering
\resizebox{0.96\textwidth}{!}{
\begin{tabular}{l|cccccccc|cccccccc}
\toprule
 & \multicolumn{8}{c|}{\textbf{UnKEBench-Loc}}& \multicolumn{8}{c}{\textbf{AKEW-Loc (CF)}} \\
 \cmidrule(lr){2-9} \cmidrule(lr){10-17} 
\textbf{Method} & \multicolumn{2}{c}{\textbf{Ori}} & \multicolumn{2}{c}{\textbf{Para}} & \multicolumn{2}{c}{\textbf{Loc}} & \multicolumn{2}{c|}{\textbf{OA}} & \multicolumn{2}{c}{\textbf{Ori}} & \multicolumn{2}{c}{\textbf{Para}} & \multicolumn{2}{c}{\textbf{Loc}} & \multicolumn{2}{c}{\textbf{OA}}\\

 \cmidrule(lr){2-3} \cmidrule(lr){4-5} \cmidrule(lr){6-7} \cmidrule(lr){8-9} \cmidrule(lr){10-11} \cmidrule(lr){12-13} \cmidrule(lr){14-15} \cmidrule(lr){16-17}
& \textbf{BS} & \textbf{RL} & \textbf{BS} & \textbf{RL} & \textbf{BS} & \textbf{RL} & \textbf{BS} & \textbf{RL} & \textbf{BS} & \textbf{RL} & \textbf{BS} & \textbf{RL} & \textbf{BS} & \textbf{RL} & \textbf{BS} & \textbf{RL} \\
\midrule
    \rowcolor{gray!30} \multicolumn{17}{c}{\textbf{Llama3-8B-Instruct}}\\
    Pre-edit & 63.18 & 23.67 & 62.73 & 23.52 & 100.00 & 100.00 & 75.30 & 49.06
        & 64.03 & 15.74 & 40.20 & 5.52 & 100.00 & 100.00 & 68.08 & 40.42 \\
    ROME & 81.76 & 45.22 & 80.47 & 42.36 & 78.49 & 47.90 & 80.24 & 45.16 
        & 83.75 & 49.68 & 57.74 & 26.34 & 79.60 & 45.47 & 73.69 & 40.50 \\ 
    MEMIT & 75.93 & 29.83 & 74.44 & 28.53 & 83.24 & 60.38 & 77.87 & 39.58 
        & 76.40 & 31.36 & 47.81 & 15.89 & 81.98 & 55.73 & 68.73 & 34.33 \\
    AlphaEdit & 73.84 & 26.84 & 72.58 & 25.91 & 84.74 & 63.13 & 77.05 & 38.63
        & 72.44 & 24.57 & 45.68 & 14.20 & 83.44 & 59.25 & 67.19 & 32.68 \\
    UnKE & 98.35 & 93.32 & 93.66 & 79.28 & 82.30 & 53.90 & 91.44 & 75.50 
        & 99.56 & 97.97 & 60.33 & 34.14 & 77.82 & 43.29 & 79.24 & 58.47 \\
    AnyEdit & 99.79 & 99.48 & 91.87 & 77.62 & 86.24 & 60.24 & \underline{92.63} & \underline{79.11} 
        & 99.99 & 99.99 & 62.72 & 43.33 & 79.24 & 43.33 & 80.65 & 62.22 \\
    AdaLoRA+ &  87.26 & 92.17& 81.17 & 76.53 & -& -& - & -
        & -& -& -&- &-& -& -&-\\
    FT-L+ & 11.63 & 7.26 & 10.16 & 6.53 & -& -& -&-
        & -& -& -&- &-& -& -&-\\
    FT-L* & 40.31 & 11.39 & 37.29 & 8.51 & -& -& -&-
        & 42.89 & 13.12 & 31.44 & 5.24 & -& -& -&-\\
    UnKE* & 98.34 & 93.33 & 93.38 & 78.42 & -& -& -&-
        & 98.62 & 96.37 & 59.62 & 32.89 & -& -& -&- \\
    AnyEdit* & 99.86 & 99.68 & 94.70 & 85.75 & -& -& -&- 
        & 99.95 & 99.98 & 64.24 & 45.31  & -& -& -&- \\
    \midrule
    AdaLoRA-UKE & 98.57 & 93.93 & 91.57 & 71.89 & 85.80 & 56.32 & 91.98 & 74.05 
        & 100.00 & 100.00 & 82.64 & 75.18 & 77.20 & 36.62 & \textbf{86.61} & \underline{70.60} \\
    FT-UKE & 99.95 & 99.97 & 99.05 & 97.07 & 81.11 & 50.04 & \textbf{93.37} & \textbf{82.36} 
        & 100.00 & 99.99 & 74.89 & 65.51 & 80.38 & 48.52 & \underline{85.09} & \textbf{71.34} \\
    \rowcolor{gray!30} \multicolumn{17}{c}{\textbf{Qwen2.5-7B-Instruct}}\\
    Pre-edit & 64.18 & 25.88 & 64.39 & 24.02 & 100.00 & 100.00 & 76.19 & 49.97
        & 65.50 & 18.24 & 44.74 & 17.29 & 100.00 & 100.00 & 70.08 & 45.18 \\
    ROME & 84.71 & 52.34 & 81.79 & 45.36 & 84.52 & 51.22 & 83.67 & 49.64 
        & 81.25 & 50.57 & 64.07 & 31.53 & 81.92 & 46.03 & 75.75 & 42.71 \\
    MEMIT & 78.19 & 38.21 & 76.62 & 34.19 & 88.12 & 61.92 & 80.98 & 44.77 
        & 76.97 & 39.03 & 56.08 & 25.69 & 86.56 & 57.87 & 73.21 & 40.86 \\
    AlphaEdit & 80.00 & 42.01 & 78.12 & 38.22 & 82.70 & 49.44 & 80.27 & 43.22
        & 80.46 & 44.43 & 57.95 & 28.16 & 82.42 & 47.28 & 73.61 & 39.96 \\
    UnKE &96.90 & 90.49 & 83.83 & 51.29 & 82.65 & 51.58 & 87.79 & 64.45 
        & 97.46 & 90.55 & 59.20 & 29.14 & 80.69 & 45.73 & 79.11 & 55.14 \\
    AnyEdit &98.75 & 96.99 & 80.94 & 51.33 & 84.36 & 53.06 & 88.01 & 67.13 
        & 99.00 & 97.59 & 57.50 & 31.90 & 82.22 & 47.37 & 79.57 & 58.95 \\
    FT-L* & 44.02 & 13.78 & 40.33 & 12.93 & -& -& -&-
        & 46.66 & 14.63 & 32.34 & 12.31 & -& -& -&-\\
    UnKE* & 96.97 & 91.01 & 89.17 & 67.00 & -& -& -&-
         & 97.34 & 90.44 & 59.29 & 29.27 -& -& -&- \\
    AnyEdit* & 99.35 & 98.82 & 94.81 & 82.60 & -& -& -&-
         & 99.63 & 98.99 & 60.78 & 32.95 -& -& -&- \\
    \midrule
    AdaLoRA-UKE & 99.97 & 99.89 & 98.68 & 94.27 & 75.94 & 39.97 & \underline{91.53} & \underline{78.05}
     & 99.99 & 100.00 & 75.19 & 60.77 & 77.40 & 42.32 & \underline{84.19} & \underline{67.69} \\
    FT-UKE & 99.97 & 99.89 & 99.08 & 97.04 & 79.02 & 41.41 & \textbf{92.69} & \textbf{79.45} 
        & 100.00 & 99.95 & 77.88 & 71.14 & 76.74 & 38.80 & \textbf{84.87} & \textbf{69.96} \\
\bottomrule
\end{tabular}
}
\caption{Knowledge editing performance with different methods. "BS" and "RL" are short for "Bert Score" and "Rouge-L" respectively. The best results are highlighted in \textbf{bold}, and the second-best results are \underline{underlined}. +: Cited from UnKE~\citep{deng2024unke}, editing Llama2-7B-Chat on UnKEBench. *: Cited from AnyEdit~\citep{jiang2025anyedit}, same experiment setup with us.}
\label{tab:main_res}
\end{table*}

%% file: table/dif_lora_ft.tex
\newcommand{\greencell}{\cellcolor{green!30}}

\begin{table*}[t]
\centering
\resizebox{0.96\textwidth}{!}{
\begin{tabular}{l|ll|cccccccc}
\toprule
    \multirow{2}{*}{\textbf{Scope}} & \multirow{2}{*}{\textbf{Layer}} & \multirow{2}{*}{\textbf{Component}} & \multicolumn{2}{c}{\textbf{Ori}} & \multicolumn{2}{c}{\textbf{Para}} & \multicolumn{2}{c}{\textbf{Loc}} & \multicolumn{2}{c}{\textbf{OA}} \\
    \cmidrule(lr){4-5} \cmidrule(lr){6-7} \cmidrule(lr){8-9} \cmidrule(lr){10-11} 
    & & & \textbf{BS} & \textbf{RL} & \textbf{BS} & \textbf{RL} & \textbf{BS} & \textbf{RL} & \textbf{BS} & \textbf{RL} \\
    \rowcolor{gray!30} \multicolumn{11}{l}{\textbf{AdaLoRA (additional parameter fine-tuning)}} \\
    \multirow{6}{*}{final prediction token} & all & \textit{q$_{proj}$} & 100.00 & 100.00 & 53.01 & 23.49 & 84.51 & 54.58 & 79.17 & 59.35 \\
    & all & \textit{q$_{proj}$, v$_{proj}$} & 99.99 & 100.00 & 80.35 & 70.15 & 78.41 & 37.82 & 86.25 & 69.32 \\ 
     & \greencell all & \greencell \textit{q$_{proj}$, k$_{proj}$, v$_{proj}$, o$_{proj}$} & \greencell 100.00 & \greencell 100.00 & \greencell 82.64 & \greencell 75.18 & \greencell 77.20 & \greencell 36.62 & \greencell \textbf{86.61} & \greencell \textbf{70.60} \\
    \cmidrule(lr){2-11}
    & single & \textit{q$_{proj}$} & 68.55 & 18.94 & 42.28 & 13.30 & 96.65 & 87.64 & 69.16 & 39.96 \\
    & single & \textit{q$_{proj}$, v$_{proj}$} & 74.18 & 26.90 & 44.51 & 15.15 & 90.45 & 67.97 & 69.71 & 36.67 \\
    & single & \textit{q$_{proj}$, k$_{proj}$, v$_{proj}$, o$_{proj}$} & 99.83 & 99.17 & 51.17 & 23.41 & 87.64 & 59.73 & \textbf{79.54} & \textbf{60.77} \\

    \rowcolor{gray!30} \multicolumn{11}{l}{\textbf{FT (direct weight fine-tuning)}} \\
    \multirow{6}{*}{final prediction token} & all & \textit{q$_{proj}$, v$_{proj}$} & 3.34 & 1.98 & 3.08 & 1.89 & 1.77 & 27.39 & 2.73 & 10.42 \\
    & all & \textit{q$_{proj}$, v$_{proj}$, down$_{proj}$} & 3.25 & 1.36 & 3.30 & 1.34 & 1.91 & 31.43 & 2.82 & 11.37\\
    & all & \textit{down$_{proj}$} & 3.36 & 1.40 & 3.36 & 1.36 & 1.98 & 32.76 & \textbf{2.90} & \textbf{11.84} \\
    \cmidrule(lr){2-11}
    & single & \textit{q$_{proj}$, v$_{proj}$} & 11.75 & 5.29 & 11.21 & 5.22 & 15.55 & 42.42 & 12.84 & 17.64 \\
    & single & \textit{q$_{proj}$, v$_{proj}$, down$_{proj}$} & 9.92 & 4.81 & 11.09 & 6.01 & 40.36 & 35.59 & 20.46 & 15.47 \\
    & single & \textit{down$_{proj}$} & 29.44 & 12.19 & 27.10 & 10.92 & 56.10 &  41.82 &  \textbf{37.55} &  \textbf{21.64} \\
    \midrule
    \multirow{6}{*}{all target tokens} & all & \textit{q$_{proj}$, v$_{proj}$} & 89.30 & 88.36 & 86.39 & 83.30 & 13.60 & 17.01 & 63.09 & 62.89 \\
    & all & \textit{q$_{proj}$, v$_{proj}$, down$_{proj}$} & 100.00 & 99.99 & 75.68 & 65.87 & 78.00 & 45.00 & \textbf{84.56} & \textbf{70.28} \\
    & all & \textit{down$_{proj}$} & 18.35 & 13.45 & 17.75 & 13.03 & 3.67 & 27.09 & 13.26 & 17.86 \\
    \cmidrule(lr){2-11}
    & single & \textit{q$_{proj}$, v$_{proj}$} & 100.00 & 99.98 & 75.66 & 65.96 & 77.86 & 44.74 & 84.51 & 70.23 \\
    & single & \textit{q$_{proj}$, v$_{proj}$, down$_{proj}$} & 100.00 & 100.00 & 68.88 & 54.35 & 81.00 & 47.17 & 83.29 & 67.17 \\
     & \greencell single & \greencell \textit{down$_{proj}$} & \greencell 100.00 & \greencell 99.99 & \greencell 74.89 & \greencell 65.51 & \greencell 80.38 & \greencell 48.52 & \greencell \textbf{85.09} & \greencell \textbf{71.34} \\
    
\bottomrule
\end{tabular}
}
\caption{Performance of FT-based methods with different factor settings on AKEW-Loc (CF). All settings apply the chat template. The best results for each group are highlighted in bold, and the settings used in \S \ref{sec:main_res} are highlighted in green (FT-UKE, AdaLoRA-UKE). The comparison for the "chat template" can be found in Appendix~\ref{app:configuration} (Table~\ref{tab:no_temp}).}

\label{tab:dif_lora_ft}
\end{table*}

%% file: table/mmlu_loc.tex
\begin{table*}[t]

\centering
\resizebox{0.96\textwidth}{!}{
\begin{tabular}{l|ccc|ccc}
\toprule
    \textbf{Edit query source} & \multicolumn{3}{c|}{\textbf{UnKEBench-Loc}} & \multicolumn{3}{c}{\textbf{AKEW-Loc (CF)}}\\
    \cmidrule(lr){1-1} \cmidrule(lr){2-4} \cmidrule(lr){5-7}
    \textbf{Loc. data source} & \multicolumn{1}{c}{\textbf{MMLU-Loc}}& \multicolumn{2}{c}{\textbf{UnKEBench-Loc}} & \multicolumn{1}{c}{\textbf{MMLU-Loc}}& \multicolumn{2}{c}{\textbf{AKEW-Loc (CF)}} \\
    \cmidrule(lr){1-1} \cmidrule(lr){2-2} \cmidrule(lr){3-4} \cmidrule(lr){5-5} \cmidrule(lr){6-7}
    \textbf{Method} & Acc & BS & RL & Acc & BS & RL \\ 
\midrule
    \rowcolor{gray!30} Pre-edit & 64.18 & 100.00 & 100.00 
        & 64.06 & 100.00 & 100.00 \\
    ROME & 63.66 (-0.52) & 78.49 (\textcolor{red}{-21.51}) & 47.90 (\textcolor{red}{-52.10})
        & 63.26 (-0.80) & 79.60 (-20.40) & 45.47 (-54.53) \\
    MEMIT & 63.96 (-0.22) & 83.24 (-16.76) & 60.38 (-39.62)
        & \textbf{63.98} (-0.08) & 81.98 (-18.02) & 55.73 (-44.27) \\
    AlphaEdit & 63.78 (-0.40) & 84.74 (-15.26) & \textbf{63.13} (-36.87)
        & 63.84 (-0.23) & \textbf{83.44} (-16.56) & \textbf{59.25} (-40.75) \\
    UnKE & 63.28 (-0.90) & 82.30 (-17.70) & 53.90 (-46.10)
        & 62.95 (-1.11) & 77.82 (-22.18) & 43.29 (-56.71) \\
    AnyEdit & 62.56 (\textcolor{red}{-1.62}) & \textbf{86.24} (-13.76) & 60.24 (-39.76)
        & 62.89 (\textcolor{red}{-1.17}) & 79.24 (-20.76) & 43.33 (-56.67) \\
    FT-UKE & \textbf{63.98} (-0.20) & 81.11 (-18.89) & 50.04 (-49.96)
        & 63.71 (-0.35) & 80.38 (-19.62) & 48.52 (-51.48) \\
    AdaLoRA-UKE & 62.92 (-1.26) & 85.80 (-14.20) & 56.32 (-43.68)
        & 63.06 (-1.01) & 77.20 (\textcolor{red}{-22.80}) & 36.62 (\textcolor{red}{-63.38}) \\

\bottomrule
\end{tabular}
}
\caption{Comparison of Locality evaluation results using MMLU-Loc and AKEW-Loc, showing the results for editing Llama3-8B-Instruct with queries from AKEW-Loc (CF). The highest values are shown in \textbf{bold}. The values in (parentheses) indicate the decrease compared to Pre-edit, with the largest decrease marked in \textcolor{red}{red}. For result on UnKEBench, please refer to Appendix~\ref{app:loc}.}
\label{tab:mmlu_loc}
\end{table*}

%% file: secs/6.conclusion.tex
\section{Conclusion}
This paper constructs two datasets UnKEBench-Loc and AKEW-Loc (CF) designed for Unstructured Knowledge Editing (UKE) from the unstructured and structured views. With three types of Locality test data, these datasets can support direct and comprehensive evaluation of UKE Locality. Besides, we outline four factors influencing FT-based methods in UKE and provide a recipe for training FT-based methods with strong performance. Our experiment results indicate that the FT-based method with the optimal setting (FT-UKE) is surprisingly strong, surpassing all the SOTA methods. 
In batch editing scenarios, FT-UKE performs strongly as well, with its advantage over SOTA methods increasing as the batch size grows, thereby expanding the average metric lead from +6.78\% to +10.80\%.
We encourage researchers to adopt our training recipe to build a strong baseline for the UKE task in future work.

%% file: secs/7.appendix.tex
\section{Dataset Construction}
\label{app:dataset}

The DPR model used in our main experiments is trained under the in-batch negative setting, where each question is paired with one additional negative document. We employ distributed training across 6 NVIDIA 1080Ti GPUs, with each GPU processing a batch size of 6, resulting in an effective total batch size of 36. The question and document encoders are jointly trained for up to 31 epochs using the Adam optimizer with a learning rate of 1e-5, a linear learning rate scheduler with warm-up, and a dropout rate of 0.1.

Following \cite{karpukhin2020dense}'s settings, We begin by using a pre-processing script to extract clean textual content from the Wikipedia dump, filtering out semi-structured elements such as tables, infoboxes, lists, and disambiguation pages. Each article is then segmented into multiple non-overlapping text blocks of approximately 100 words, which are treated as individual retrieval documents. This process results in roughly 21M documents in total. To construct the locality dataset, we first employ the trained DPR model to retrieve high similarity documents for each question. For the RelDoc setting, we use the Stanford OpenIE toolkit to extract triples from the unstructured facts and ensure that none of the extracted entity-relation combinations appear in the retrieved documents. RandDoc involves randomly sampling documents from the entire corpus, while explicitly excluding those that appear in the top-100 retrieval results. StructTrip is constructed by sampling questions from the structured editing dataset KnowEdit \citep{zhang2024comprehensive}, with additional filtering to guarantee that the involved entities do not reoccur in the retrieved documents. For the final statistics of each Locality test,  we use the \textit{word\_tokenize} function from the NLTK \citep{bird-loper-2004-nltk} library to count the number of tokens.

\section{Experiment Details}
\label{app:exp_details}
The settings for FT-based methods and ROME, are primarily based on those used in EasyEdit \citep{wang2023easyedit}, while all other configurations follow the original implementation of AnyEdit \citep{jiang2025anyedit} to ensure consistency. All experiments are conducted on a single NVIDIA H20 GPU with 96GB of memory. The following are their important hyperparameter configuration contents.

\paragraph{UnKE} UNKE performs edits at layer 7. The model is trained with a learning rate of $5 \times 10^{-1}$ for 25 optimization steps, using a weight attenuation coefficient of $1 \times 10^{-3}$. This is followed by 50 additional optimization steps with a reduced learning rate of $2 \times 10^{-4}$ to further refine the parameter updates.

\paragraph{AnyEdit} For Llama3-8B-Instruct, the standard AnyEdit configuration is adopted, where editing is performed at layer 7 with a clamp norm factor of 4. The fact token is defined as the last token in the prompt. During optimization, all parameters within both the attention and MLP layers are updated. A learning rate of $2 \times 10^{-4}$ is used for 50 gradient steps. For key-value representation updates, 25 optimization steps are conducted with a higher learning rate of 0.5. The loss is applied at layer 31, and a weight decay of $1 \times 10^{-3}$ is employed. To mitigate unintended knowledge drift, 20 samples are drawn from the original model distribution to serve as constraints. For chunked editing, a chunk size of 40 tokens is used without overlap. For Qwen2.5-7B-Instruct, the configuration remains the same, except that the loss is applied at layer 27 and the chunk size is increased to 50 tokens.

\paragraph{ROME and MEMIT} 
The key difference between ROME and MEMIT lies in the number of layers involved in the editing process. ROME performs updates exclusively on layer 5, whereas MEMIT operates on a broader range of layers: [4, 5, 6, 7, 8]. Both methods are optimized using 25 steps with a learning rate of $5 \times 10^{-1}$, a weight attenuation coefficient of $1 \times 10^{-3}$, and a KL regularization factor of 0.0625.

\paragraph{FT} FT updates the model weights with a learning rate of $5 \times 10^{-4}$, performing 25 optimization steps for each training sample. The update is restricted to a single transformer layer, and we explore two optimization objectives: \textbf{prompt-last}, which supervises the representation at the last token of the prompt, and \textbf{target-new}, which directly supervises the representation of the injected target entity.

\paragraph{AdaLoRA} 
For the AdaLoRA experiments, we adopt parameter-efficient tuning by inserting low-rank adapter modules into all transformer layers. We set the LoRA rank to 8, the scaling factor \texttt{lora\_alpha} to 32, and apply a dropout rate of 0.1. The learning rate is set to $5 \times 10^{-3}$. The target modules include the attention projections: \textit{q$_{proj}$}, \textit{k$_{proj}$}, \textit{v$_{proj}$}, and \textit{o$_{proj}$}.

\section{Configuration Details}
\label{app:configuration}

\paragraph{Component Selection} Each Transformer layer primarily consists of two submodules: the multi-head self-attention (MHSA) module and the feed-forward network (FFN) module. In the MHSA module, the input hidden states are projected through linear layers to produce the query (\textit{q$_{proj}$}), key (\textit{k$_{proj}$}), and value (\textit{v$_{proj}$}) vectors. These are used to compute attention scores and aggregate contextual information, followed by an output projection (\textit{o$_{proj}$}) that maps the result back to the original hidden dimension. In the FFN module, the hidden representations are transformed through a nonlinear activation and projected back using the down projection (\textit{down$_{proj}$}) layer.

\paragraph{Impact of Chat Template.} 
We also investigate the impact of chat template adaptation on editing performance. Our results show a clear performance gap between models with and without chat template alignment. For instance, on the AdaLoRA setting, using the template yields an average BS of 86.61\% and  RL of 70.60\%, compared to 71.24\% and 56.34\% without the template, a relative improvement of 14.26\% in RL. Similarly, for FT, the RL score improves from 43.22\% (w/o template) to 71.34\% (w. template), a dramatic gain of over 28.12\%.  These results highlight the importance of aligning with the model’s expected input format. Omitting the chat template leads to suboptimal edits, likely due to mismatches in prompt structure and internal representations. Therefore, template adaptation should be considered a necessary step for effective knowledge editing, especially when working with instruction-tuned models.

\input{table/no_temp}

\section{Performance of Batch Ediging}
\label{app:batch}

\input{table/batch_res} 
As shown in Table~\ref{tab:batch_res}, we further evaluate the performance of various editing methods under different batch sizes (1, 10, 50, 100) on the AKEW-Loc dataset. FT-UKE consistently demonstrates strong and stable performance across all batch sizes, maintaining high factual accuracy (Ori) while effectively preserving both generalization (Para) and locality (Loc). Notably, its advantages become increasingly evident as the batch size grows. While methods like MEMIT exhibit relatively stable behavior, most notably, AdaLoRA-UKE, whose accuracy drops rapidly with increasing batch size. 
In contrast, FT-UKE maintains a well-balanced performance, leading to a clear overall advantage (OA) over competing approaches.

\section{Locality Results}
\label{app:loc}
Table~\ref{tab:loc_res} provides a detailed comparison of three distinct types of locality, \textit{RelDoc}, \textit{RandDoc}, and \textit{StructTrip}, and illustrates how different editing methods perform under these settings on the \textsc{UnKEBench-Loc} and \textsc{AKEW-Loc (CF)} datasets.

\input{table/loc_res}

%% file: table/no_temp.tex
\begin{table}[t]

\centering
\resizebox{0.46\textwidth}{!}{
\begin{tabular}{l|cccccc|cc}
\toprule
\textbf{Method} & \multicolumn{2}{c}{\textbf{Ori}} & \multicolumn{2}{c}{\textbf{Para}} & \multicolumn{2}{c}{\textbf{Loc}} & \multicolumn{2}{|c}{\textbf{OA}} \\
 \cmidrule(lr){2-3} \cmidrule(lr){4-5} \cmidrule(lr){6-7} \cmidrule(lr){8-9}
 & \textbf{BS} & \textbf{RL} & \textbf{BS} & \textbf{RL} & \textbf{BS} & \textbf{RL} & \textbf{BS} & \textbf{RL} \\
\midrule
    \rowcolor{gray!30} \multicolumn{9}{l}{AdaLora (additional parameter fine-tuning)} \\ 
    \hspace{1em}w. template& \textbf{100.00} & \textbf{100.00} & \textbf{82.64} & \textbf{75.18} & \textbf{77.20} & 36.62 & \textbf{86.61} & \textbf{70.60} \\
    \hspace{1em}w/o. template & 89.33 & 95.07 & 48.40 & 33.14 & 76.00 & \textbf{40.80} & 71.24 & 56.34 \\
    \rowcolor{gray!30} \multicolumn{9}{l}{FT (direct weight fine-tuning)} \\
    \hspace{1em}w. template & \textbf{100.00} & \textbf{99.99} & 74.89 & 65.51 & 80.38 & 48.52 & \textbf{85.09} & \textbf{71.34} \\
    \hspace{1em}w/o. template & 80.00 & 42.01 & \textbf{78.12} & \textbf{38.22} & \textbf{82.70} & \textbf{49.44} & 80.27 & 43.22 \\

\bottomrule
\end{tabular}
}
\caption{Results of editing Llama3-8B-Instruct with (w.) and with out(w/o.) chat template on AKEW-Loc (CF). Settings of other factors keep same with the best setting in Table~\ref{tab:dif_lora_ft}.}
\label{tab:no_temp}
\end{table}

%% file: table/batch_res.tex
\begin{table}[t]

\centering
\resizebox{0.46\textwidth}{!}{
\begin{tabular}{l|cccccccc}
\toprule
\textbf{Method} & \multicolumn{2}{c}{\textbf{Ori}} & \multicolumn{2}{c}{\textbf{Para}} & \multicolumn{2}{c}{\textbf{Loc}} & \multicolumn{2}{c}{\textbf{OA}}\\

 \cmidrule(lr){2-3} \cmidrule(lr){4-5} \cmidrule(lr){6-7} \cmidrule(lr){8-9}
& \textbf{BS} & \textbf{RL} & \textbf{BS} & \textbf{RL} & \textbf{BS} & \textbf{RL} & \textbf{BS} & \textbf{RL}  \\
\midrule
    \rowcolor{gray!30} \multicolumn{9}{c}{\textbf{Batch Size=1}}\\
    ROME & 83.75 & 49.68 & 57.74 & 26.34 & 79.60 & 45.47 & 73.69 & 40.50 \\
    MEMIT & 76.40 & 31.36 & 47.81 & 15.89 & 81.98 & 55.73 & 68.73 & 34.33 \\
    UNKE & 99.56 & 97.97 & 60.33 & 34.14 & 77.82 & 43.29 & 79.24 & 58.47 \\
    AnyEdit & 99.99 & 99.99 & 62.72 & 43.33 & 79.24 & 43.33 & 80.65 & 62.22 \\
    AdaLoRA-UKE & 100.00 & 100.00 & 82.64 & 75.18 & 77.20 & 36.62 & 86.61 & 70.60\\
    FT-UKE & 100.00 & 99.99 & 74.89 & 65.51 & 80.38 & 48.52 & 85.09 & 71.34 \\
    
    \rowcolor{gray!30} \multicolumn{9}{c}{\textbf{Batch Size=10}} \\
    ROME & 72.86 & 28.13 & 51.60 & 19.52 & 79.55 & 45.02 & 68.00 & 30.89 \\
    MEMIT & 54.15 & 14.23 & 40.18 & 12.21 & 72.28 & 47.34 & 55.53 & 24.59 \\
    UNKE & 99.61 & 98.31 & 57.27 & 32.69 & 73.43 & 34.76 & 76.77 & 55.25 \\
    AnyEdit & 99.86 & 99.78 & 56.75 & 39.02 & 78.16 & 36.37 & 78.25 & 58.39 \\
    AdaLoRA-UKE & 90.17 & 67.61 & 53.60 & 26.52 & 80.80 & 41.34 & 74.86 & 45.16 \\
    FT-UKE & 99.90 & 99.76 & 75.53 & 63.03 & 76.42 & 37.57 & 83.95 & 66.79 \\

    \rowcolor{gray!30} \multicolumn{9}{c}{\textbf{Batch Size=50}} \\
    ROME & 71.41 & 23.71 & 49.88 & 17.48 & 80.31 & 48.89 & 67.20 & 30.03 \\
    MEMIT & 67.89 & 18.51 & 43.44 & 13.33 & 95.39 & 85.04 & 68.91 & 38.96 \\
    UNKE & 99.57 & 97.90 & 54.70 & 28.51 & 76.03 & 36.93 & 76.76 & 54.45 \\
    AnyEdit & 75.44 & 48.76 & 51.10 & 31.84 & 79.45 & 37.17 & 68.66 & 39.26 \\
    AdaLoRA-UKE & 77.74 & 47.43 & 52.05 & 29.10 & 66.91 & 29.94 & 65.56 & 35.49 \\
    FT-UKE & 99.91 & 99.72 & 68.74 & 50.27 & 77.29 & 36.80 & 81.98 & 62.26 \\
    
    \rowcolor{gray!30} \multicolumn{9}{c}{\textbf{Batch Size=100}}\\
    ROME & 72.36 & 24.95 & 49.44 & 17.86 & 80.55 & 48.57 & 67.45 & 30.46 \\
    MEMIT & 68.27 & 18.69 & 42.24 & 13.18 & 96.39 & 88.05 & 68.97 & 39.98 \\
    UNKE & 92.54 & 75.38 & 53.15 & 25.58 & 77.43 & 40.54 & 74.37 & 47.17 \\
    AnyEdit & 71.33 & 45.34 & 50.51 & 31.40 & 80.01 & 37.77 & 67.28 & 38.17 \\
    AdaLoRA-UKE & 42.96 & 23.50 & 31.12 & 18.04 & 38.89 & 24.83 & 37.66 & 22.12 \\ 
    FT-UKE & 99.96 & 99.73 & 68.24 & 49.16 & 76.41 & 35.92 & 81.54 & 61.60 \\

\bottomrule
\end{tabular}
}
\caption{Detailed results of Figure \ref{fig:batch_res}: Editing Llama3-8B-Instruct on AKEW-Loc (CF) with batch size of 1, 10, 50, 100.}
\label{tab:batch_res}
\end{table}

%% file: table/loc_res.tex
\begin{table*}[t]

\centering
\resizebox{0.86\textwidth}{!}{
\begin{tabular}{l|cccccc|cccccc}
\toprule
  & \multicolumn{6}{c|}{\textbf{UnKEBench-Loc}}& \multicolumn{6}{c}{\textbf{AKEW-Loc (CF)}} \\
  \cmidrule(lr){2-7} \cmidrule(lr){8-13} 

 \textbf{Method} & \multicolumn{2}{c}{\textbf{RelDoc}} & \multicolumn{2}{c}{\textbf{RandDoc}} & \multicolumn{2}{c|}{\textbf{StructTrip}} & 
    \multicolumn{2}{c}{\textbf{RelDoc}} & \multicolumn{2}{c}{\textbf{RandDoc}} & \multicolumn{2}{c}{\textbf{StructTrip}} \\

 \cmidrule(lr){2-3} \cmidrule(lr){4-5} \cmidrule(lr){6-7} \cmidrule(lr){8-9} \cmidrule(lr){10-11} \cmidrule(lr){12-13}  
& \textbf{BS} & \textbf{RL} & \textbf{BS} & \textbf{RL} & \textbf{BS} & \textbf{RL} & \textbf{BS} & \textbf{RL} & \textbf{BS} & \textbf{RL} & \textbf{BS} & \textbf{RL} \\
\midrule
    \rowcolor{gray!30} \multicolumn{13}{c}{\textbf{Llama3-8B-Instruct}}\\
    ROME & 76.64 & 47.80 & 79.12 & 50.78 & 79.71 & 45.13 
        & 80.76 & 49.63 & 79.77 & 48.87 & 79.60 & 45.47 \\
    MEMIT & 80.61 & 56.90 & 83.23 & 59.70 & 85.88 & 64.54 
        & 82.41 & 57.36 & 81.70 & 56.54 & 81.81 & 53.30 \\
    AlphaEdit & 82.43 & 58.96 & 84.49 & 61.72 & 87.30 & 68.69
        & 83.79 & 58.19 & 82.86 & 58.78 & 83.66 & 60.79 \\
    UNKE & 80.37 & 51.69 & 83.46 & 55.49 & 83.08 & 54.51 
        & 80.31 & 51.57 & 79.72 & 50.24 & 73.43 & 28.07 \\
    AnyEdit & 83.41 & 52.71 & 86.24 & 59.43 & 89.06 & 68.58 
        & 81.31 & 48.38 & 80.94 & 49.95 & 75.47 & 31.67 \\
    AdaLoRA & 75.67 & 40.48 & 78.86 & 44.01 & 78.44 & 36.36 
        & 78.88 & 40.98 & 78.93 & 43.08 & 77.42 & 29.39 \\
    FT-M & 72.16 & 45.02 & 77.70 & 55.23 & 83.05 & 58.67 
        & 76.40 & 45.29 & 82.24 & 56.00 & 85.78 & 60.48 \\

    \rowcolor{gray!30} \multicolumn{13}{c}{\textbf{Qwen2.5-7B-Instruct}}\\
    ROME & 86.17 & 51.72 & 87.06 & 53.58 & 80.31 & 48.36 
        & 83.61 & 46.00 & 84.61 & 47.73 & 77.53 & 44.36 \\
    MEMIT & 89.11 & 62.71 & 90.94 & 64.79 & 84.33 & 58.27 
        & 88.21 & 58.35 & 90.02 & 61.83 & 81.45 & 53.42 \\
    AlphaEdit & 83.99 & 49.37 & 86.45 & 52.16 & 77.65 & 46.80
        & 84.76 & 47.54 & 85.98 & 50.73 & 76.51 & 43.55 \\
    UNKE & 84.47 & 52.74 & 86.05 & 55.10 & 77.43 & 46.90 
        & 82.75 & 45.30 & 84.44 & 48.21 & 74.86 & 43.67 \\
    AnyEdit & 86.07 & 54.43 & 87.90 & 57.89 & 79.10 & 46.85 
        & 84.48 & 47.30 & 86.08 & 51.72 & 76.10 & 43.08 \\
    AdaLoRA & 76.48 & 41.47 & 77.92 & 43.01 & 71.39 & 34.79 
        & 79.48 & 43.53 & 80.79 & 46.76 & 70.69 & 40.07 \\
    FT-M & 79.02 & 41.60 & 83.57 & 46.04 & 72.52 & 36.59 
        & 76.17 & 37.11 & 83.18 & 44.69 & 70.87 & 34.58 \\
\bottomrule
\end{tabular}
}
\caption{Locality of post-edit models across three types.}
\label{tab:loc_res}
\end{table*}